\pdfoutput=1

\documentclass[10pt,twocolumn,letterpaper]{article}

\usepackage[letterpaper,margin=0.8in,columnsep=0.3in]{geometry}
\usepackage{times}
\usepackage[dvipsnames]{xcolor}
\usepackage{graphicx}
\usepackage{amsmath}
\usepackage{amssymb}
\usepackage{booktabs}
\usepackage[numbers,sort&compress]{natbib}
\usepackage[format=plain,labelformat=simple,labelsep=period,font=small]{caption}
\usepackage[hyphens]{url}
\usepackage[shortlabels,inline]{enumitem}
\setlist[itemize]{noitemsep,leftmargin=*,topsep=0em}
\setlist[enumerate]{noitemsep,leftmargin=*,topsep=0em}

\usepackage{microtype}
\usepackage{amsfonts}
\usepackage{array}
\usepackage{algorithm}
\usepackage{algpseudocode}
\usepackage{flafter}
\usepackage{dblfloatfix}

\graphicspath{{fig/}}

\newcommand{\dG}{\Delta G}
\newcommand{\dGbar}{\Delta\bar{G}}
\newcommand{\cond}{\mathbf{c}}
\newcommand{\best}[1]{\textbf{\boldmath #1}}

\definecolor{linkblue}{rgb}{0.21,0.49,0.74}
\usepackage[pagebackref,breaklinks,colorlinks,allcolors=linkblue]{hyperref}
\usepackage[capitalize]{cleveref}
\crefname{section}{Sec.}{Secs.}
\Crefname{section}{Section}{Sections}
\Crefname{table}{Table}{Tables}
\crefname{table}{Tab.}{Tabs.}

\title{AquaProbe: Efficient Adaptation of Frozen Vision Models for Underwater Images}

\author{Md Shihabul Islam Shovo$^{1}$ \quad
Nishi Kanta Paul$^{1}$ \quad
Trung Dong$^{2}$ \quad
Xiaomin Lin$^{2}$\\
$^{1}$NIMISHES Lab, Dhaka, Bangladesh \quad
$^{2}$University of South Florida\\
{\tt\small \{shihabul900, nishikantapaul108\}@gmail.com \quad \{dongt, xlin2\}@usf.edu}
}
\date{}

\begin{document}
\maketitle
\begin{abstract}
Reliable visual perception is essential for safe autonomous underwater
exploration. Yet vision models pretrained on terrestrial images become less
reliable underwater, where turbidity, backscatter, and wavelength-dependent
attenuation alter scene appearance. Existing adaptation methods either spread adapters equally across
the encoder depth or select layers using the divergence between two fixed
domains. In this paper, we introduce AquaProbe for measuring how the location of the
domain gap changes with water type and imaging range. AquaProbe separates
newly introduced divergence from that accumulated through a frozen vision
encoder using paired clean and physically rendered images and allocates a
fixed LoRA budget according to the measured profile. The segmentation mean Intersection-over-Union (mIoU) for the proportional LoRA
rank allocation, based on the measured profile, is 70.1; this is a better
result than uniform LoRA rank allocation by 1.9 mIoU points while having the
same parameter budget of 0.59 million adapter parameters. The three-way
router, conditioned on estimated condition clusters, achieves 70.6 mIoU on the
standard SUIM split and 63.4 AUDRC under held-out degradation.
\end{abstract}

\section{Introduction}
\label{sec:intro}

Safe autonomous underwater exploration requires reliable visual perception,
which is often built on vision backbones pretrained on data from terrestrial
environments. Underwater images are distributed differently from the
pretraining distribution due to physically caused degradation: light is
attenuated at wavelength-dependent rates, and part of the observed radiance is
backscattered ambient light rather than scene radiance
\citep{akkaynak2018revised}. With increased turbidity and imaging
range, contrast and visibility decrease, particular colour channels collapse,
and high frequency details are lost, so the domain shift is condition dependent.

Fine-tuning such a pretrained encoder is difficult because underwater
annotations are scarce and computation and energy resources for onboard
inference are constrained; consequently, full fine-tuning is rarely attractive.
Parameter-efficient fine-tuning helps address these problems by freezing the
backbone and training only a small set of additional weights
\citep{houlsby2019adapters,hu2022lora,jia2022vpt}. This reduces adaptation
cost, but given a limited budget for allocation, it is essential to determine
how this budget should be divided between the layers of the encoder.

Adapter allocation is commonly uniform, based on the assumption that the
domain shift affects all depths equally. Adaptive algorithms, including
importance-driven budget allocation \citep{zhang2023adalora}, use the
optimisation statistics collected during training as opposed to directly
measuring the shift. Layer-wise similarity analysis provides such
a measurement \citep{kornblith2019cka,raghu2021vitcnn} and has been used to set
rank before training \citep{chu2026repsam}. However, this analysis is performed
between a fixed pair of domains and therefore yields a static adapter
allocation, whereas underwater conditions vary continuously with water
type and imaging range, so one static profile may be insufficient. To measure
the variation and allocate LoRA rank accordingly, we introduce AquaProbe.

With AquaProbe's degradation-conditioned representational divergence
analysis, we address three problems with the measurements of the
representation gap. First, the representation gap between a pristine set and
a corrupted one is characteristic only of this pair of distributions
\citep{bendavid2010theory} and cannot indicate whether its location reflects
the encoder or imaging conditions. Since underwater degradation has a
closed-form model, we render clean images with a two-term physical model
parametrized by Jerlov water type and imaging range, so that the profile is a
function of these parameters and the stability of the gap is an empirical
question. Second, layer-wise divergence is reported cumulatively and compounds
with depth, so the curve rises regardless of where the damage starts.
Instead, the per-block increase in divergence within the frozen encoder
is recorded. Third, allocation based on representational
measurements assumes that a large gap warrants more capacity, but this
assumption is rarely validated against downstream behaviour. A single-block
sweep characterizes this relationship. Finally, we derive a
rank-based allocation method from these profiles and compare it with uniform,
random, reversed, and prior allocation methods, as well as an
exploratory router operating over nested prefixes of per-cluster allocations.

Our contributions are as follows:
\begin{itemize}
  \setlength{\itemsep}{0pt}
  \item We present a degradation-conditioned probe that measures the
        per-block increase in divergence and show that it behaves
        differently from the commonly reported cumulative divergence.
  \item The profile drift index is compared to the average from a null distribution
        obtained through resampling, with peak dispersion greater than one
        block in two out of the three backbones and falling below that
        threshold for the third.
  \item We investigate the relationship between divergence and adaptation
        utility through a single-block sweep, converting an assumption made by
        allocation-by-analysis methods into a measured quantity.
  \item A rank allocation policy is developed from the profile, and its performance is
        compared to uniform, random, reversed, and two prior allocation
        methods under a matched budget, along with an exploratory router using nested
        per-cluster prefixes with a matching parameter count ($0.03\%$).
\end{itemize}

\section{Related Work}
\label{sec:related}

\subsection{Parameter-Efficient Adaptation}
Adapter layers \citep{houlsby2019adapters}, prompt tuning
\citep{jia2022vpt}, parallel bottleneck modules \citep{chen2022adaptformer}, and low-rank
weight updates \citep{hu2022lora} enable efficient fine-tuning of a frozen
backbone by learning a few auxiliary parameters. Improvements mostly use
optimisation signals and include AdaLoRA \citep{zhang2023adalora}, which prunes
low-importance singular directions during training, DyLoRA
\citep{valipour2023dylora}, which jointly trains a range of ranks, and DoRA
\citep{liu2024dora}, which separates magnitude from direction. I-LoRA
\citep{li2026ilora} also scores parameter importance with integrated
gradients for adaptive rank allocation. Closest to our work is RepSAM
\citep{chu2026repsam}, which uses centred kernel alignment between the
pretraining distribution and a target robotic dataset to divide 32 SAM
encoder layers into three fixed-rank bands ahead of fine-tuning.
A parallel line of work investigates which layers to adapt rather than how
much capacity each should receive \citep{ma2026layerselect,zhao2026rank1}. Dense
prediction under water is addressed by ViT-UWA \citep{jia2026vituwa}, which
utilises a high-frequency prior and a convolutional detail branch along with a
pretrained transformer without measuring capacity needs.

In all cases, capacity is set by design, from optimisation statistics, or
from a fixed domain pair, without modelling how allocation needs change with
physical underwater conditions. In the proposed method, divergence is computed in relation to the
physical attributes of the shift rather than between two fixed datasets, so
that a field is created instead of a single scalar quantity and stationarity
can be tested rather than assumed. It also differences the profile across
depth instead of reading it cumulatively and tests whether larger gaps warrant
more capacity through downstream utility.

\subsection{Layer-Wise Analysis of Representations}
Feature transferability depends on depth \citep{yosinski2014transferable}.
Canonical correlation methods \citep{raghu2017svcca,morcos2018pwcca}, together
with centred kernel alignment \citep{kornblith2019cka}, measure representations
across networks and depths. Such methods have been used to compare
vision transformers to convolutional networks \citep{raghu2021vitcnn}. The reliability of centred
kernel alignment has been examined because CKA can be sensitive to outliers
and high-variance directions \citep{davari2023ckareliability}. The results
are usually presented for a particular pair of conditions and cumulatively at
each depth, even for capacity allocation \citep{chu2026repsam}. The cumulative
curve, however, cannot cleanly locate new divergence due to the propagation of
initial perturbations through later blocks. Instead, we analyse the
condition-dependent changes per block using centred kernel alignment as the
main statistic, a mutual nearest-neighbour profile for robustness, and the
per-block difference to localise divergences.

\subsection{Underwater Image Formation and Perception Data}
Robustness is commonly evaluated using synthetic corruptions at
multiple severity levels \citep{hendrycks2019imagenetc}, while degradation in the
underwater setting has physically meaningful parameters: appearance is modelled as a
distance-attenuated direct transmission term with an additional additive
backscatter term \citep{akkaynak2018revised}. Restoration
methods invert these effects using transmission estimates, haze-line priors,
or known scene range
\citep{drews2013udcp,berman2017hazelines,akkaynak2019seathru}, whereas
learning-based enhancement methods train on paired or unpaired data
\citep{li2020uieb,li2021ucolor,islam2020funiegan,peng2023ushape}. Available
evaluation data span segmentation \citep{islam2020suim,lian2023watermask},
detection \citep{fu2023ruod,liu2021duo}, marine debris
\citep{hong2020trashcan}, raw-reference enhancement pairs
\citep{li2020uieb}, and colour-chart-calibrated imagery
\citep{berman2021squid}. UWCNN \citep{anwar2020uwcnn} also applies the
same physical model for synthesising ten Jerlov types of underwater images
from indoor RGB-D scenes. This allows for a structural approach with controlled
axes of measurement: following the same procedure, we manipulate water type
and imaging range to analyse representation drift and reuse the rendered
conditions for shared training augmentation and router supervision.

\section{Method}
\label{sec:method}

\subsection{Overview}
The process comprises two consecutive stages. A probe locates where the
measured gap grows in a frozen vision transformer
\citep{dosovitskiy2021vit} $f_\theta$ over a grid of physical conditions, and
an adaptation stage turns the resulting profile into a per-block LoRA
rank budget, optionally with a router when the profile depends on imaging
conditions. We write $F^{(\ell)}(I)\in\mathbb{R}^{N\times d}$ for the token
features after block $\ell$, with $N$ patch tokens of width $d$, and $\ell=0$
for the patch-embedding output.

\begin{figure*}[!t]
  \centering
  \includegraphics[width=0.71\textwidth]{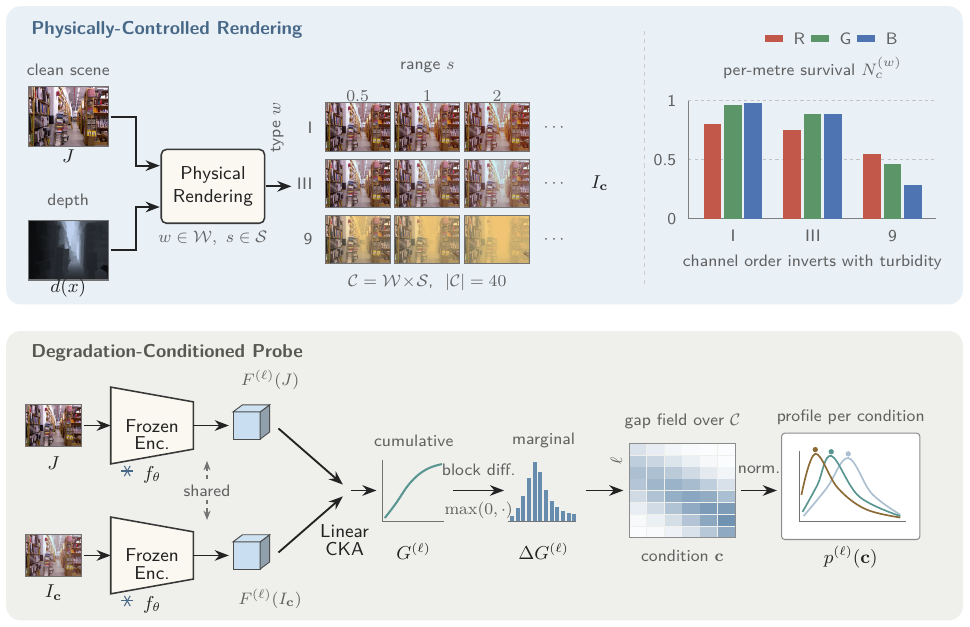}
  \caption{Clean RGB-D scenes are rendered across Jerlov water types and imaging
  ranges, and clean and rendered pairs pass through the same frozen encoder to
  give the per-block divergence field and the normalised profile.}
  \label{fig:probe}
\end{figure*}

\begin{figure*}[!t]
  \centering
  \includegraphics[width=0.75\textwidth]{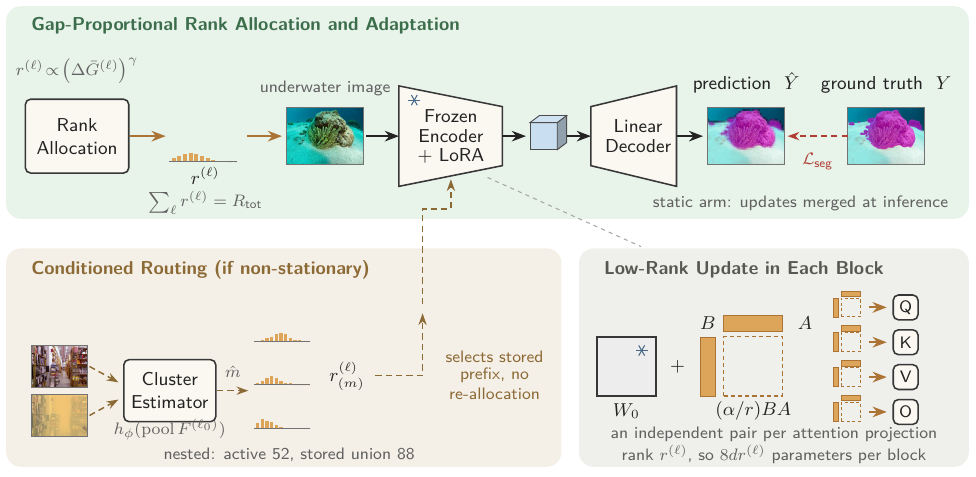}
  \caption{The measured profile sets each block's LoRA rank under a fixed
  budget, the rank-allocated encoder feeds a linear decoder, and a router
  selects among nested per-cluster prefixes when the profile is condition
  dependent.}
  \label{fig:adaptation}
\end{figure*}

\subsection{Physically Controlled Rendering}
\label{sec:render}
Clean and rendered pairs enter the probe in \cref{fig:probe}, whose
profile drives the adaptation stage in \cref{fig:adaptation}. Rendering uses
the simplified two-term formation model adopted for underwater image
synthesis \citep{anwar2020uwcnn}. For clean scene radiance $J$ and
camera-to-scene range $z(x)$ at pixel $x$, the observed image is
\begin{equation}
  I_c(x) = J_c(x)\,t_c(x) + B_c^{\infty}\bigl(1-t_c(x)\bigr),
  \qquad c\in\{R,G,B\},
  \label{eq:formation}
\end{equation}
where $B_c^{\infty}$ denotes the veiling light. The transmission
$t_c(x)=(N_c^{(w)})^{z(x)}$ is controlled by the per-metre survival fraction
$N_c^{(w)}\in(0,1)$ of channel $c$ for Jerlov water type $w$, or equivalently
by the attenuation coefficient $\beta_c^{(w)}=-\ln N_c^{(w)}$. We use the ten
tabulated types \citep{anwar2020uwcnn} of the standard Jerlov
classification \citep{williamson2022jerlov}: five open-ocean types (I, IA, IB,
II, III) and five increasingly turbid coastal types (1C, 3C, 5C, 7C,
9C). The table carries a signature that the probe should reproduce: in clear
oceanic water blue survives best and red is extinguished first, an ordering
reversed in the most turbid coastal type.

Range is an explicit second axis rather than a nuisance variable. With
$\hat{d}(x)\in[0,1]$ the min-max normalised source depth map and
$s\in\mathcal{S}$ a range multiplier,
\begin{equation}
  z(x) = s\bigl[z_{\min} + (z_{\max}-z_{\min})\,\hat{d}(x)\bigr],
  \label{eq:range}
\end{equation}
and a condition is the pair
$\cond=(w,s)\in\mathcal{C}=\mathcal{W}\times\mathcal{S}$.
As \cref{eq:formation} is a per-pixel photometric map, labels transfer
unchanged from the source data and the measurement needs no additional
annotation.

\subsection{Degradation-Conditioned Probe}
\label{sec:probe}
Agreement between clean and rendered features is measured using linear centred
kernel alignment \citep{kornblith2019cka}. For centred feature matrices $X$ and
$Y$, with samples represented by rows,
\begin{equation}
  \mathrm{CKA}(X,Y) =
  \frac{\lVert Y^{\top}X\rVert_F^{2}}
       {\lVert X^{\top}X\rVert_F\,\lVert Y^{\top}Y\rVert_F}.
  \label{eq:cka}
\end{equation}
Patch tokens are mean-pooled to obtain one image per row, which stabilises the
estimate. The cumulative gap at depth $\ell$ under condition $\cond$ is
\begin{equation}
  G^{(\ell)}(\cond) = 1-\mathrm{CKA}\!
  \left(F^{(\ell)}(J),\,F^{(\ell)}(I_{\cond})\right),
  \label{eq:cumulative}
\end{equation}
evaluated on a fixed image pool while holding scene content constant so that
only the degradation varies.

\Cref{eq:cumulative} entangles two effects: a perturbation introduced
early propagates through later blocks, so $G^{(\ell)}$ tends to increase with
depth regardless of where the perturbation originated. The quantity of
interest for allocation is instead the increment local to block $\ell$,
\begin{equation}
  \dG^{(\ell)}(\cond) =
  \max\!\left(0,\; G^{(\ell)}(\cond)-G^{(\ell-1)}(\cond)\right),
  \label{eq:marginal}
\end{equation}
where rectification handles small negative values caused by estimation noise;
it alters $3.4\%$ of block-condition pairs, all in the final four blocks,
where the increment is close to zero. Alignment is computed in a different
feature space at every depth, so the two terms in \cref{eq:marginal} are not
strictly commensurable. The difference thus marks where the recorded
divergence increases, which we define as newly introduced divergence. It makes
no distinction between new and propagated corruption and is analysed as an
allocation indicator in \cref{sec:utility}. Dividing by
$\sum_j \dG^{(j)}(\cond)$ yields the normalised profile $p^{(\ell)}(\cond)$,
which captures where the gap concentrates, not its magnitude, which turbid
water raises at every depth.

The same construction attributes the gap to individual colour channels:
restoring one channel of the rendered image to its clean value, leaving the
other two degraded, and recomputing $\sum_\ell \dG^{(\ell)}$ defines the
attribution $A_c(\cond)$ as the resulting reduction in total added divergence.
The tabulated survival fractions predict which channel should carry more of
the degradation for each water type, so $A_c$ is a falsifiable check rather
than a descriptive statistic.

\subsection{Stationarity of the Profile}
\label{sec:stationarity}
The profile drift index is defined as the mean pairwise Jensen--Shannon divergence
\citep{lin1991jsd} over the condition grid,
\begin{equation}
  \mathrm{PDI} =
  \frac{2}{|\mathcal{C}|(|\mathcal{C}|-1)}
  \sum_{i<j}\mathrm{JSD}\!\left(p(\cond_i)\,\Vert\,p(\cond_j)\right).
  \label{eq:pdi}
\end{equation}
This index is meaningful only relative to a null distribution of estimation
noise, which we construct by resampling the full image pool under a single
fixed condition and recomputing~\cref{eq:pdi}. The profile is declared
non-stationary when the observed index exceeds the upper $5\%$ of the null and
the dispersion of the peak block
$\ell^{*}(\cond)=\arg\max_\ell \dG^{(\ell)}(\cond)$ exceeds one block. We also
report the Spearman correlation between $\ell^{*}$ and each physical axis,
because drift that tracks a physical parameter is more informative than
unstructured variation.

\begin{algorithm}[!t]
\caption{Rank allocation by water-filling projection}
\label{alg:alloc}
\small
\begin{algorithmic}[1]
\Require profile $\dGbar\in\mathbb{R}_{>0}^{L}$, budget $R_{\text{tot}}$,
         exponent $\gamma$, bounds $r_{\min},r_{\max}$
         with $L\,r_{\min}\le R_{\text{tot}}\le L\,r_{\max}$
\Ensure integer ranks $r$ with $\sum_\ell r^{(\ell)} = R_{\text{tot}}$
\State $w \gets (\dGbar)^{\gamma}$
\State bisect for the level $\tau>0$ with
       $\sum_{\ell}\operatorname{clip}(\tau w^{(\ell)},r_{\min},r_{\max})
       = R_{\text{tot}}$
\State $\tilde{r}^{(\ell)} \gets \operatorname{clip}(\tau w^{(\ell)},
       r_{\min},r_{\max})$ for all $\ell$
\State assign $\lfloor \tilde{r}^{(\ell)}\rfloor$ to each block
\State distribute the remaining $R_{\text{tot}}-\sum_{\ell}
       \lfloor\tilde{r}^{(\ell)}\rfloor$ units to the largest fractional
       remainders
\State \Return $r$
\end{algorithmic}
\end{algorithm}

\subsection{Per-Block Adaptation Utility}
\label{sec:utility}
Allocation based on the measurement assumes that a block with greater
divergence benefits more from additional capacity; we measure this relationship
rather than assume it. A LoRA module of small fixed rank $r_{\text{probe}}$ is
attached to one block at a time, with all other blocks frozen and the same
decoder and schedule in every run. The utility $U^{(\ell)}$ is the mIoU
improvement over the frozen encoder, and $\rho_{\text{valid}}$ is the Spearman
correlation between the condition-averaged increment $\dGbar^{(\ell)}$ and
$U^{(\ell)}$. Because adapters may interact, single-block utility is not the
marginal contribution of a block within a joint allocation, so this statistic
validates the allocation rule without establishing its optimality.

\subsection{Gap-Proportional Rank Allocation}
\label{sec:allocation}
Given a budget $R_{\text{tot}}$, a sharpening exponent $\gamma\ge 0$, and rank
bounds $[r_{\min},r_{\max}]$, the target allocation assigns ranks in proportion
to the condition-averaged divergence raised to the sharpening exponent:
\begin{equation}
  \tilde{r}^{(\ell)} = R_{\text{tot}}
  \frac{\bigl(\dGbar^{(\ell)}\bigr)^{\gamma}}
       {\sum_{j}\bigl(\dGbar^{(j)}\bigr)^{\gamma}} .
  \label{eq:alloc}
\end{equation}
The exponent interpolates between uniform allocation at $\gamma=0$, which
provides a correctness check because it must reproduce the uniform baseline
exactly, and a winner-take-all allocation as $\gamma$ increases.
A water-filling projection maps the continuous target onto the feasible
integer set, applying clipping before integerisation because reversing these
operations would violate the budget constraint.

LoRA is used as published \citep{hu2022lora}: it transforms a frozen weight
$W_0$ into
$W_0 + (\alpha/r)BA$ with $A\in\mathbb{R}^{r\times d}$ normal and
$B\in\mathbb{R}^{d\times r}$ zero. Applying an independent pair to each of the
four attention projections of every block gives
\begin{equation}
  P_{\text{adapter}}
  = \sum_{\ell=1}^{L} 8\,d\,r^{(\ell)}
  = 8\,d\,R_{\text{tot}} ,
  \label{eq:budget}
\end{equation}
so equal-budget static LoRA methods have identical counts.

\subsection{Conditioned Routing}
\label{sec:routing}
Where the measured profile varies substantially across imaging conditions
(\cref{sec:stationarity}), a single allocation may not match the full input
distribution. We partition the condition grid into $M$ clusters through
agglomerative clustering with average linkage, applied directly to the
pairwise Jensen--Shannon distance matrix, the square root of each
pairwise divergence in \cref{eq:pdi}; using pairwise distances avoids a
Euclidean centroid of the profiles. Applying the procedure of
\cref{sec:allocation} (\cref{alg:alloc}) within each cluster yields $M$
allocations $r^{(\ell)}_{(m)}$.

Independent adapters would multiply the parameter count by $M$, making the
equal-budget comparison invalid. Allocations instead share nested factors,
similar to DyLoRA's jointly trained rank range \citep{valipour2023dylora}:
block $\ell$ has one pair of factors at rank $\max_m r^{(\ell)}_{(m)}$, and
cluster $m$ uses the first $r^{(\ell)}_{(m)}$ columns of $B$ and rows of $A$
with scaling $\alpha/r^{(\ell)}_{(m)}$. The stored budget, $\sum_{\ell}\max_m
r^{(\ell)}_{(m)}$, is counted by \cref{eq:budget}; the per-cluster
budget is chosen so that the union meets the static arm's $R_{\text{tot}}$.
Each cluster is therefore active at a lower rank than the static allocation.

A two-layer perceptron $h_\phi$ takes pooled features from an early block
$\ell_0$, taken from an adapter-free pass, and predicts cluster
membership under the cross-entropy loss $\mathcal{L}_{\text{route}}$,
supervised on rendered data whose cluster label is known by construction. At
inference, the predicted cluster selects the prefix of the stored factors, so
ranks are not recomputed. The parameters of $h_\phi$ are included in the
reported budget and $R_{\text{tot}}$ is reduced accordingly, preserving the
matched comparison.

\subsection{Training and Inference}
\label{sec:training}
A single $1\times1$ convolution followed by bilinear upsampling converts patch
tokens to class logits; shared by all arms, this decoder keeps the encoder
representation on the critical path. The loss is $\mathcal{L}=\mathcal{L}_{\text{CE}}
+\lambda_{\text{dice}}\mathcal{L}_{\text{Dice}}
+\lambda_{\text{route}}\mathcal{L}_{\text{route}}$, where
$\lambda_{\text{route}}=0$ unless the router is on. At inference, the backbone
is frozen and the static arm merges the low-rank update into $W_0$ without an
additional branch. The routed arm's input-dependent prefix prevents merging,
so it maintains the low-rank branch and recomputes the first $\ell_0$ of
$L$ blocks after routing.

\section{Experimental Setup}
\label{sec:experiments}

\begin{figure*}[!t]
  \centering
  \includegraphics[width=0.86\textwidth]{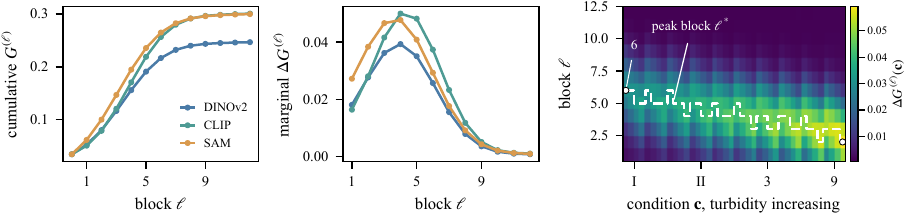}
  \caption{Condition-averaged cumulative divergence saturates with depth while
condition-averaged marginal divergence is unimodal and early. Over the condition grid the peak of the marginal field,
  traced by the dashed line, moves towards earlier blocks as turbidity rises.}
  \label{fig:gapoverview}
\end{figure*}

\subsection{Datasets and Preprocessing}
Rendering uses the 1449 densely annotated RGB-D pairs of the labelled NYU
Depth~V2 subset \citep{silberman2012nyu}, which supplies the metric depth that
\cref{eq:range} requires and has previously been used for underwater scene
synthesis \citep{anwar2020uwcnn}. Each scene is rendered under all ten Jerlov
types and four range multipliers $\mathcal{S}=\{0.5,1,2,4\}$, producing
$|\mathcal{C}|=40$ conditions with a probe pool of $n=1449$ pairs each.

The primary downstream task is semantic segmentation on SUIM
\citep{islam2020suim}, with 1525 training/validation images, 110 test
images, and eight classes. The test split is held out from probing,
allocation, and all hyperparameter selection; the sweep over $\gamma$ uses a
held-out slice of the training split. TrashCan \citep{hong2020trashcan}, whose
instance masks are collapsed into 16 material classes, is a second benchmark
captured in deep water under artificial lighting, and the 890 raw images with
paired references in UIEB \citep{li2020uieb} support probing on genuine
photographs.

Images are resized so that the shorter side matches the encoder resolution and
normalised with the backbone's statistics; masks use nearest-neighbour
interpolation. During adaptation, the renderer augments real SUIM training
images identically in every arm. SUIM and TrashCan lack depth, so $z(x)$ is
an inverted monocular inverse-depth estimate \citep{ranftl2022midas}
passed through the normalisation of \cref{eq:range}. These images are already
attenuated, so the water model is applied a second time and the scene is not
calibrated; as this is identical in every arm and for every robustness test
image, it shifts the operating point without confounding the ranking.

\subsection{Evaluation Metrics}
\label{sec:metrics}
The measurement stage reports $G^{(\ell)}$, $\dG^{(\ell)}$, the drift index
and its ratio to the resampling null, and the dispersion of $\ell^{*}$ and its
Spearman correlation with each physical axis; mutual $k$-nearest-neighbour
alignment at $k=10$ is a robustness check.

The adaptation stage reports mean intersection over union (mIoU), boundary
F-score, and trainable parameters. Robustness to conditions not used for
allocation or training is the area under the degradation robustness curve
(AUDRC): the trapezoidal integral of mIoU over SUIM test images re-rendered at
held-out range multipliers $\mathcal{S}'=\{0.25,0.75,1.5,3,5\}$, normalised by
the interval width, because a simple mean would overweight the unevenly spaced
short-range end. $\mathcal{S}'$ is disjoint from $\mathcal{S}$, whereas the
ten water types are shared, so AUDRC holds out imaging range, not water type.

\subsection{Backbones and Baselines}
We probe DINOv2 ViT-B/14 \citep{oquab2024dinov2}, CLIP ViT-B/16
\citep{radford2021clip}, and the SAM ViT-B image encoder
\citep{kirillov2023sam}. DINOv2 serves the main adaptation comparison,
with equal-budget analyses on SAM and TrashCan using the profile measured for
each backbone.

Static LoRA arms share a budget of $R_{\text{tot}}=96$, or rank eight per
block under uniform allocation, which is $0.59$M adapter parameters and
$0.68\%$ of the backbone; the shared $6.2$K linear decoder is trained in every
arm but excluded from this count. The comparison includes a frozen encoder
with a decoder only, uniform LoRA, random allocation over five seeds, the
reversed profile, AdaLoRA \citep{zhang2023adalora}, and a banded allocation
following RepSAM \citep{chu2026repsam} that groups cumulative alignment
between a terrestrial and a SUIM pool into three depth bands. We also evaluate
the proposed static and routed variants and full fine-tuning as an
unequal-budget upper reference. The reversed arm is a directional control: if
a peaked allocation helped for reasons unrelated to the measured profile,
reversing it would not reduce accuracy.

\subsection{Implementation Details}
Experiments use PyTorch on Google Colab Pro with an NVIDIA A100 40\,GB
accelerator. LoRA adapts every block's query, key, value, and output
projections with $\alpha=16$ and, for non-uniform allocations, bounds
$r_{\min}=0$ and $r_{\max}=24$, allowing blocks to be skipped; the held-out
training slice selects $\gamma=1$. AdamW \citep{loshchilov2019adamw} uses
learning rates $3\times10^{-4}$ for adapters and $1\times10^{-3}$ for the
decoder, weight decay $0.01$, a cosine schedule with one warm-up epoch, batch
size 16, 40 epochs, and $\lambda_{\text{dice}}=1$. The router uses $M=3$
clusters, pooled features from $\ell_0=2$, and hidden width 64, yielding
$768\times64+64\times3=49{,}344$ parameters; its three allocations operate at
rank 52 over a stored union of 88. Main adaptation methods use three seeds and
the random baseline five. The main comparison reports mIoU as mean and
standard deviation; boundary F-score, AUDRC, and ablation mIoUs have no
reported standard deviations.

\begin{table}[!t]
\centering
\caption{Each backbone is summarised over the 40-condition grid, with
PDI\,/\,null the ratio of the observed index to the mean of the resampling
null.}
\label{tab:stationarity}
\small
\setlength{\tabcolsep}{3.5pt}
\begin{tabular}{lccccc}
\toprule
Backbone & $\ell^{*}$ & $\mathrm{std}(\ell^{*})$ & PDI &
PDI\,/\,null & $\rho(w,\ell^{*})$ \\
\midrule
DINOv2 ViT-B/14 & 2--6 & 1.07 & 0.039 & 5.2  & $-0.86$ \\
CLIP ViT-B/16   & 2--7 & 1.22 & 0.070 & 10.2 & $-0.88$ \\
SAM ViT-B       & 2--6 & 0.98 & 0.030 & 4.1  & $-0.86$ \\
\bottomrule
\end{tabular}
\end{table}

Each similarity estimate uses the complete pool of $n=1449$ images,
which exceeds the feature width of 768, rather than disjoint
subsamples, to maximise estimator stability. Error bars come from 20
bootstrap resamples at full sample size, the null uses 200 bootstrap profiles,
and the utility sweep uses $r_{\text{probe}}=4$ with two seeds per block. The
stationarity test and utility sweep were frozen before any allocation arm was
trained, so the allocation was not tuned to the comparison on which it is
evaluated. Probing takes about nine accelerator-minutes and the utility sweep
about six accelerator-hours; the latter validates the principle once and is
not part of applying the method.

\section{Results and Discussion}
\label{sec:results}

\subsection{Layer-Wise Localization of the Domain Gap}
The rendered conditions show different degrees of agreement with
real-image feature statistics. Fr\'echet Inception Distance (FID)
\citep{heusel2017fid} is minimised by coastal type~1C for SUIM ($38.6$ versus
$52.4$ for the next closest type) and type~3C for UIEB ($44.1$ versus $57.9$).
\Cref{sec:realprobe}
examines transfer to real imagery.

\begin{table*}[!t]
\centering
\caption{Equal-budget comparison on SUIM with DINOv2 ViT-B/14. Trainable parameters
exclude the shared decoder; the routed count includes the router and matches the
static arms within approximately $0.03\%$. Full fine-tuning is an unequal-budget reference.}
\label{tab:main}
\small
\setlength{\tabcolsep}{4pt}
\begin{tabular}{llcccccc}
\toprule
& Method & Allocation signal &
\begin{tabular}[b]{@{}c@{}}Trainable\\params (M)\end{tabular} &
\begin{tabular}[b]{@{}c@{}}Trainable /\\backbone (\%)\end{tabular} &
mIoU (\%) & Bound.\ F (\%) & AUDRC (\%) \\
\midrule
B0  & Frozen encoder with decoder & none                 & 0.00  & 0.00 & $61.4\pm0.3$ & 44.1 & 52.1 \\
B2  & Uniform LoRA                & none                 & 0.59  & 0.68 & $68.2\pm0.5$ & 51.3 & 59.8 \\
B5a & Random allocation           & none                 & 0.59  & 0.68 & $68.0\pm0.9$ & 51.0 & 59.4 \\
B5b & Reversed profile            & reversed $\dGbar^{(\ell)}$ & 0.59 & 0.68 & $66.9\pm0.6$ & 49.8 & 58.1 \\
B6a & AdaLoRA \citep{zhang2023adalora} & training importance & 0.59 & 0.68 & $68.9\pm0.5$ & 52.1 & 60.3 \\
B6b & RepSAM bands \citep{chu2026repsam} & cross-domain $G^{(\ell)}$ & 0.59 & 0.68 & $69.3\pm0.5$ & 52.8 & 60.9 \\
B7  & AquaProbe, static           & $\dGbar^{(\ell)}$    & 0.59  & 0.68 & $70.1\pm0.4$ & 53.6 & 61.9 \\
B8  & AquaProbe, routed           & per-cluster          & 0.59  & 0.68 & \best{$70.6\pm0.5$} & \best{54.2} & \best{63.4} \\
\midrule
B1  & Full fine-tuning            & none                 & 86.60 & 100  & $71.8\pm0.7$ & 55.4 & 63.9 \\
\bottomrule
\end{tabular}
\end{table*}

\begin{figure*}[!t]
  \centering
  \includegraphics[width=0.77\textwidth]{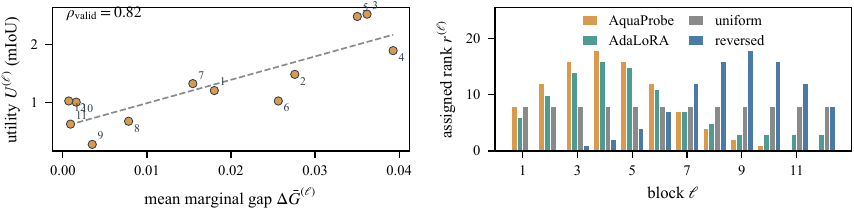}
  \caption{Single-block adaptation utility rises with the measured increment,
  with points labelled by block index, and the four strategies compared at the
  shared budget $R_{\text{tot}}=96$ differ in where they place capacity.}
  \label{fig:validity}
\end{figure*}
\Cref{fig:gapoverview} contrasts the two measurements. For all three
backbones, the cumulative curve saturates near the final blocks, so
$\arg\max_\ell G^{(\ell)}$ always places the gap at the network output,
whereas the condition-averaged marginal curve is unimodal and peaks between
the third and fifth blocks.

\subsection{The Gap Moves with the Water}
\Cref{fig:gapoverview} shows the peak shifting from block six in clear,
short-range water to block two in the most turbid, long-range condition;
\cref{tab:stationarity} quantifies this drift. The drift index of DINOv2 is
$5.2$ times the null mean, and its peak dispersion exceeds one block. CLIP
drifts more than DINOv2, and SAM, although it narrowly misses the dispersion
criterion, has a drift index several times the null mean. These summaries
indicate profile drift but do not establish formal rejection under
\cref{sec:stationarity} without the upper-tail null thresholds.

Drift is structured: correlation with turbidity is strongly negative for every
backbone, versus approximately $-0.42$ for range. Mutual
$k$-nearest-neighbour and alignment profiles agree with Spearman correlations
between $0.86$ and $0.91$, supporting robustness to the choice of statistic.

Channel attribution reproduces the rendering signature. In clear type~I water,
red restoration dominates, $A_R=0.171\,[0.146,0.198]$ versus
$A_B=0.031\,[0.019,0.044]$; in turbid type~9C, the order reverses,
$A_B=0.312\,[0.276,0.349]$ versus $A_R=0.198\,[0.170,0.228]$: the probe
captures the sign reversal implied by Jerlov attenuation, not generic
contrast loss.

On 890 unrendered UIEB raw-reference pairs, the profile correlates with the
synthetic profiles at $\rho=0.79$; both it and FID select type~3C
at unit range. Grouping pairs by UCIQE
\citep{yang2015uciqe}, a quality proxy for turbidity, reproduces the drift,
with peak blocks four, three, and two. Each group has about 297 samples of
768-wide features, near the reduced-pool regime below where profile
correlation drops to $0.71$, so the ordering is qualitative.\label{sec:realprobe}
As UIEB references are enhancement outputs, not measured radiance, this
corroborates real-image transfer without providing ground truth.

\subsection{Divergence Predicts Adaptation Utility}
\label{sec:validity}
\Cref{fig:validity} relates the measured increment to the utility of
adapting one block at a time. The rank correlation is
$\rho_{\text{valid}}=0.82$, with a bootstrap interval of $[0.47,0.95]$, so
blocks with larger increments benefit most from a small amount of adaptation
capacity. Twelve blocks form a small sample, so the interval is wide: utility
peaks at the third block and the largest increment at the fourth. The
cumulative gap gives $-0.71$, ranking the
late blocks highest while the utility sweep ranks them lowest, so
\cref{eq:marginal} changes which blocks are selected, not merely the magnitude
of the signal.

This does not conflict with RepSAM \citep{chu2026repsam}, which measures the
same statistic cumulatively between two fixed datasets, finds it lowest in
shallow layers, and concentrates rank there. Measurements across datasets
confound degradation with content mismatch, which is largest where features
are least abstract. Holding content fixed removes this term, leaving a
monotonic cumulative curve whose location is uninformative, so differencing is
required in our setting.

\subsection{Equal-Budget Comparison}
\Cref{tab:main} presents the main comparison, and the right panel of
\cref{fig:validity} shows each allocation. The measured profile
concentrates rank on blocks two to six and leaves the two deepest blocks
unadapted, outperforming uniform allocation by $1.9$ mIoU at an identical
parameter count. Random allocation matches uniform allocation within noise,
whereas the reversed control is worse than both, which isolates the
directional signal in the profile from a generic benefit of peaked allocation.
The prior allocation baselines lie in between: AdaLoRA improves by
$0.7$ mIoU and the banded RepSAM allocation by $1.1$ mIoU, because its
cross-domain measurement also places rank early. The remaining $0.8$ gap
reflects per-block resolution and a degradation-matched measurement.

Routing adds $0.5$ on the standard split, with overlapping mean $\pm$ standard
deviation intervals for the static and routed arms. Its larger gain appears
under shift, but AUDRC has no seed variance and no shuffled-route control
isolates condition-aware selection, so routing remains exploratory. Its
classifier achieves $91.3\%$ accuracy on held-out rendered conditions. Full
fine-tuning remains ahead but trains about 147 times as many
parameters, excluding the shared decoder.

In our sweep, uniform allocation reaches $70.2$ only at $R_{\text{tot}}=384$
(\cref{tab:ablation}), which requires four times the adapter parameters. Under
held-out AUDRC ranges all arms degrade, but routed and static allocation remain
$3.6$ and $2.1$ points above uniform; the static allocation
becomes mismatched as conditions depart from its condition-averaged profile.

\begin{table}[!t]
\centering
\caption{All values are mIoU. Static LoRA settings use $R_{\text{tot}}=96$,
except the uniform-budget sweep (``ours'' uses 96). Routed counts include the
router and match within approximately $0.03\%$; full fine-tuning uses an
unequal budget. Bold marks selected settings; the SAM row is unmarked because
seed standard deviations are not reported.}
\label{tab:ablation}
\small
\setlength{\tabcolsep}{2.6pt}
\begin{tabular}{lccccc}
\toprule
Sharpening $\gamma$ & 0 & 0.5 & 1 & 2 & 4 \\
mIoU (\%)           & 68.2 & 69.4 & \best{70.1} & 69.7 & 68.9 \\
\midrule
Allocation signal   & \multicolumn{2}{c}{marginal $\dGbar^{(\ell)}$}
                    & \best{70.1}
                    & \multicolumn{2}{c}{cumulative $68.4$} \\
Similarity metric   & \multicolumn{2}{c}{linear CKA} & \best{70.1}
                    & \multicolumn{2}{c}{mutual $k$-NN $69.9$} \\
Lower rank bound    & \multicolumn{2}{c}{$r_{\min}=0$} & \best{70.1}
                    & \multicolumn{2}{c}{$r_{\min}=1$: $69.8$} \\
\midrule
Uniform $R_{\text{tot}}$ & 96 & 192 & 288 & 384 & ours \\
mIoU (\%)                & 68.2 & 69.4 & 69.9 & 70.2 & \best{70.1} \\
\midrule
Transfer            & Unif. & AdaLoRA & Static & Routed & Full \\
SUIM, SAM           & 66.4 & 67.0 & 67.9 & 68.0 & 69.6 \\
TrashCan            & 44.1 & 44.6 & 45.6 & \best{45.9} & 47.0 \\
\bottomrule
\end{tabular}
\end{table}

\Cref{tab:ablation} isolates the individual design choices. The sharpening
exponent performs best at $\gamma=1$, and $\gamma=0$ reproduces uniform
allocation exactly, confirming that \cref{alg:alloc} is consistent with the
uniform baseline. Driving allocation with the content-matched cumulative gap
costs $1.7$ points, leaving only $0.2$ above uniform, because the cumulative
curve saturates late.

Measurement choices leave the rank ordering of the profile intact: the class
token instead of mean pooling, a halved condition grid, and a range interval
widened to $0.5$--$8$\,m give correlations of $0.94$, $0.97$, and $0.91$ with
the default profile. Reducing the image pool to 256 lowers this to $0.71$, the
sample-size effect anticipated in \cref{sec:experiments} and the one
measurement choice with a substantial effect.

\subsection{Generalisation, Robustness and Ablations}
\label{sec:transfer}
The lower block of \cref{tab:ablation} repeats the equal-budget comparison
outside the primary setting. On SAM ViT-B, allocation from that backbone's own
profile improves by $1.5$ mIoU over uniform, compared with $1.9$ on DINOv2.
SAM falls below the peak-dispersion criterion of \cref{sec:stationarity}, so
it is not routed by default; forcing routing raises mIoU from $67.9$ to
$68.0$, and without seed standard deviations we cannot conclude whether
routing helps. TrashCan improves by $1.5$ under the same rule, although its
deep-sea footage under artificial lighting is not described by
\cref{eq:formation}, so the profile is transferred to conditions under which
it was not measured.

\subsection{Limitations}
The probe uses indoor RGB-D scenes because metric depth is scarce underwater.
Unlike the revised model \citep{akkaynak2018revised}, \cref{eq:formation}
has a common coefficient for direct and backscattered light, so the rendered
images approximate rather than calibrate water conditions. The utility sweep
cannot capture adapter interactions, and adaptation covers two backbones, two
datasets, and only semantic segmentation. The router's labels, its $91.3\%$
accuracy, and the robustness test share one renderer, with only range held
out (\cref{sec:metrics}); its margin supports conditioning under controlled
range shift, not deployment shift, and water outside the Jerlov table is
uncharacterised.

\section{Conclusion}
\label{sec:conclusion}

We detect the blocks at which the underwater domain gap becomes wider
within a frozen vision encoder and distribute a fixed adaptation budget
accordingly. Cumulative divergence
pinpoints the gap at the network output, whereas per-block divergence reaches
early peaks that track adaptation utility. Under rendered degradation,
peak locations vary with turbidity for all three backbones, with dispersion
greater than one block for DINOv2 and CLIP. At an equal budget, profile-based
allocation gains $1.9$ mIoU on SUIM with DINOv2, $1.5$ on SUIM with SAM, and
$1.5$ on TrashCan with DINOv2; uniform allocation needs four times the adapter
parameters to match it in our SUIM/DINOv2 sweep. Data are public; code
will accompany the paper.

{
    \small
    \bibliographystyle{ieeenat_fullname}
    \bibliography{main}
}

\end{document}